\documentclass[runningheads]{llncs}
\usepackage{amsmath,graphicx}
\usepackage{textcomp} 
\usepackage{xspace} 
\usepackage{tikz}
\usepackage{xcolor}
\usepackage{marvosym}
\usepackage{tabularx}
\usepackage{dirtytalk}
\usepackage{float}

\makeatletter
\DeclareRobustCommand\onedot{\futurelet\@let@token\@onedot}
\def\@onedot{\ifx\@let@token.\else.\null\fi\xspace}
\def\eg{\emph{e.g}\onedot}

\def\etal{\emph{et~al}\onedot}

\newcommand{\NoiseVoid}{\mbox{\textsc{Noise2Void}}\xspace}
\newcommand{\DenoiSeg}{\mbox{\textsc{DenoiSeg}}\xspace}

\newcommand{\UNet}{\mbox{\textsc{U-Net}}\xspace}
\newcommand{\img}{\boldsymbol{x}}
\newcommand{\seg}{\boldsymbol{y}}

\usepackage{array}
\newcolumntype{L}[1]{>{\raggedright\let\newline\\\arraybackslash\hspace{0pt}}m{#1}}
\newcolumntype{C}[1]{>{\centering\let\newline\\\arraybackslash\hspace{0pt}}m{#1}}
\newcolumntype{R}[1]{>{\raggedleft\let\newline\\\arraybackslash\hspace{0pt}}m{#1}}

\newcommand\blfootnote[1]{%
  \begingroup
  \renewcommand\thefootnote{}\footnote{#1}%
  \addtocounter{footnote}{-1}%
  \endgroup
}

\newcommand\figTeaser{
\begin{figure}[t]
    \centering
    \includegraphics[width=.8\linewidth]{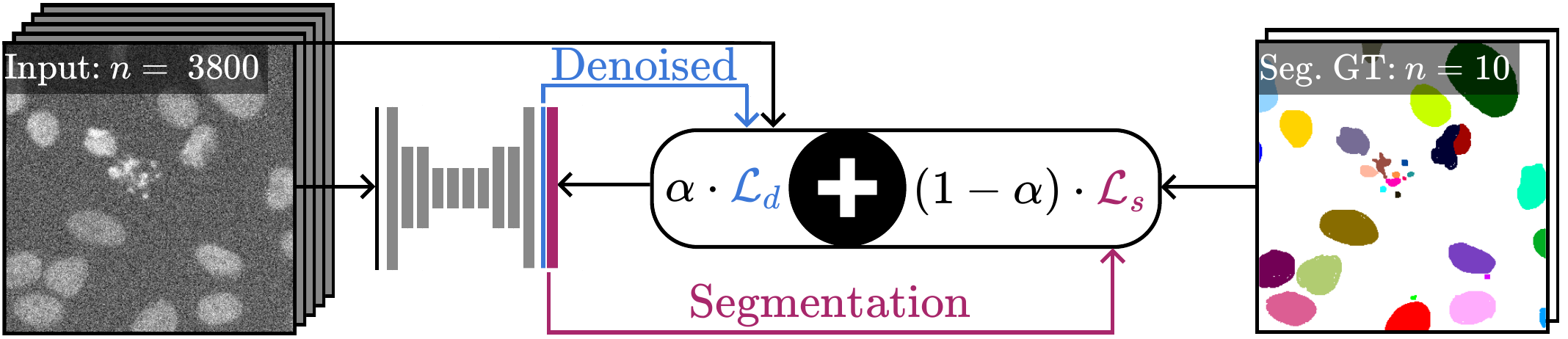}

    \caption{The proposed \DenoiSeg training scheme. 
    A \UNet is trained with a joint self-supervised denoising loss~($\mathcal{L}_d$) and a classical segmentation loss~($\mathcal{L}_s$). 
    Both losses are weighted with respect to each other by a hyperparameter $\alpha$. In this example, $\mathcal{L}_d$ can be computed on all $3800$ training
    patches, while $\mathcal{L}_s$ can only be computed on the $10$ available annotated ground truth patches that are available for segmentation.
    }
    
    \label{fig:teaser}
\end{figure}
}

\newcommand\figDSB{
\begin{figure}[t]
    \centering
    \begin{minipage}{.02\linewidth}
    \begin{tikzpicture}
            \draw (0, 0) node[rotate=90] {\textcolor{white}{wsp}DSB n20};
        \end{tikzpicture}
    \end{minipage}
    \begin{minipage}{.97\linewidth}
        \includegraphics[width=.49\linewidth,trim={0.6cm 1.3cm 0.6cm 0.5cm},clip]{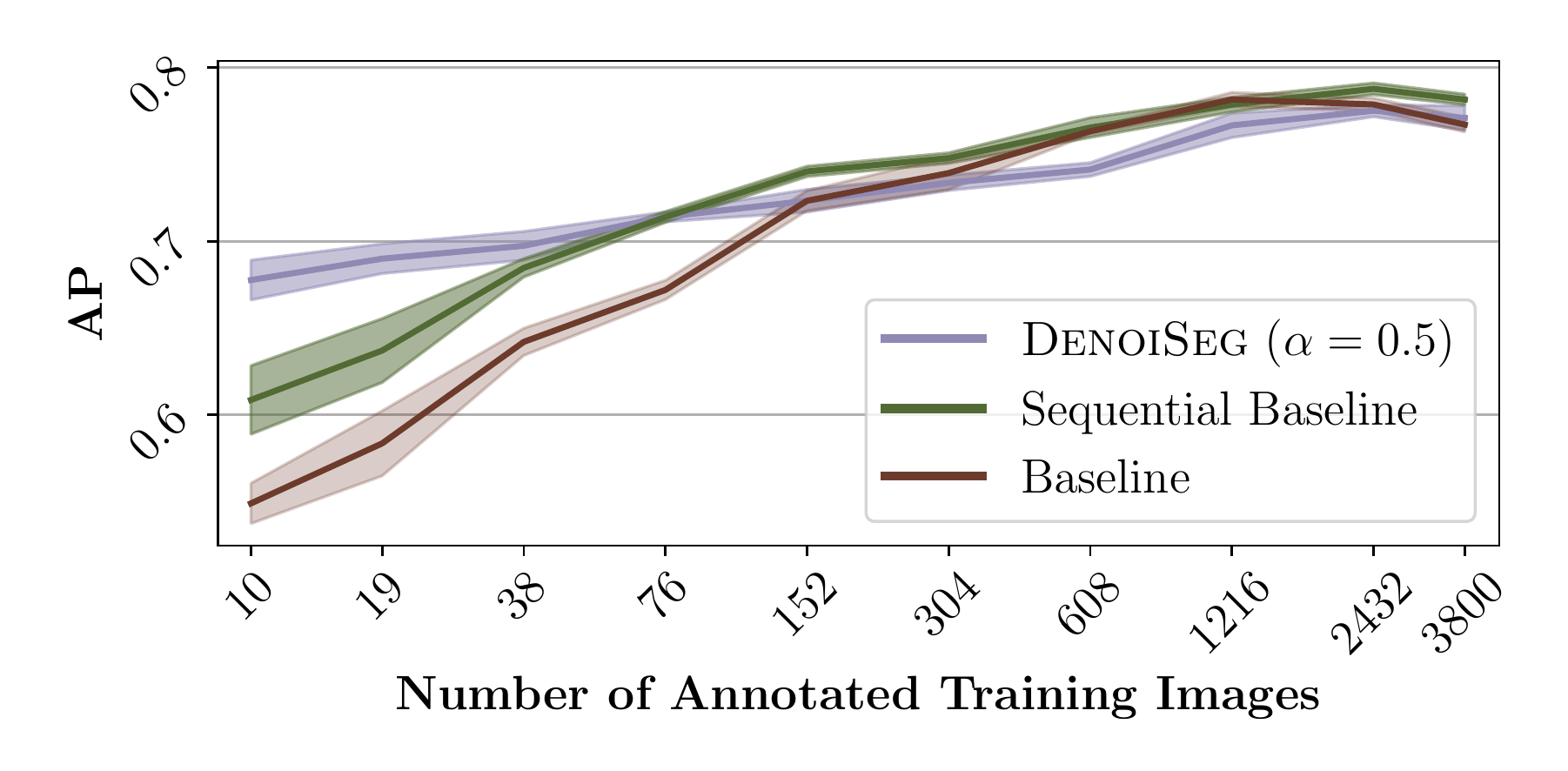}
        \includegraphics[width=.49\linewidth,trim={0.6cm 1.3cm 0.6cm 0.5cm},clip]{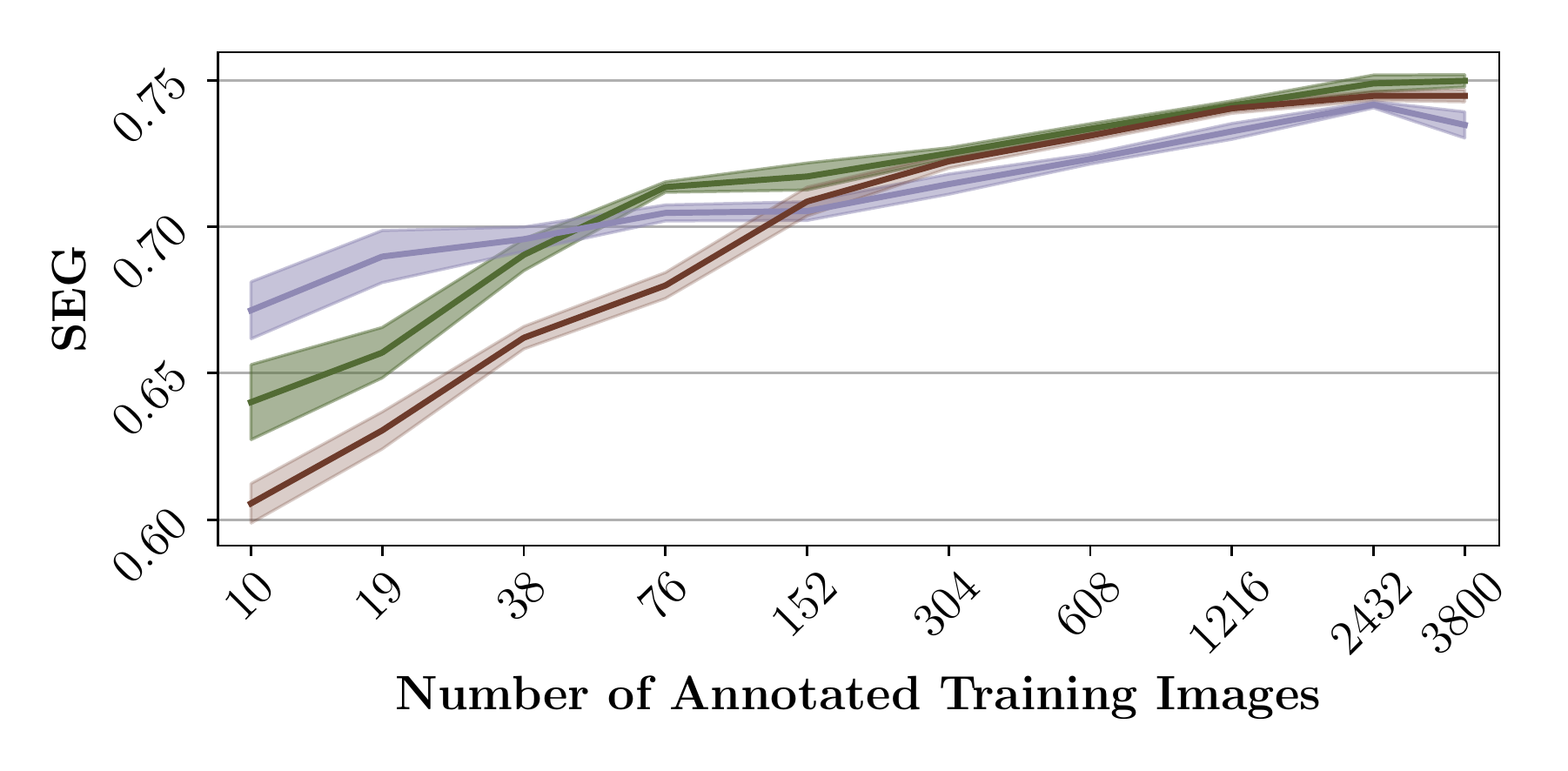}
    \end{minipage}

    \begin{minipage}{.02\linewidth}
    \begin{tikzpicture}
            \draw (0, 0) node[rotate=90] {\textcolor{white}{ws}DSB n10};
        \end{tikzpicture}
    \end{minipage}
    \begin{minipage}{.97\linewidth}
        \includegraphics[width=.49\linewidth,trim={0.6cm 1.3cm 0.6cm 0.5cm},clip]{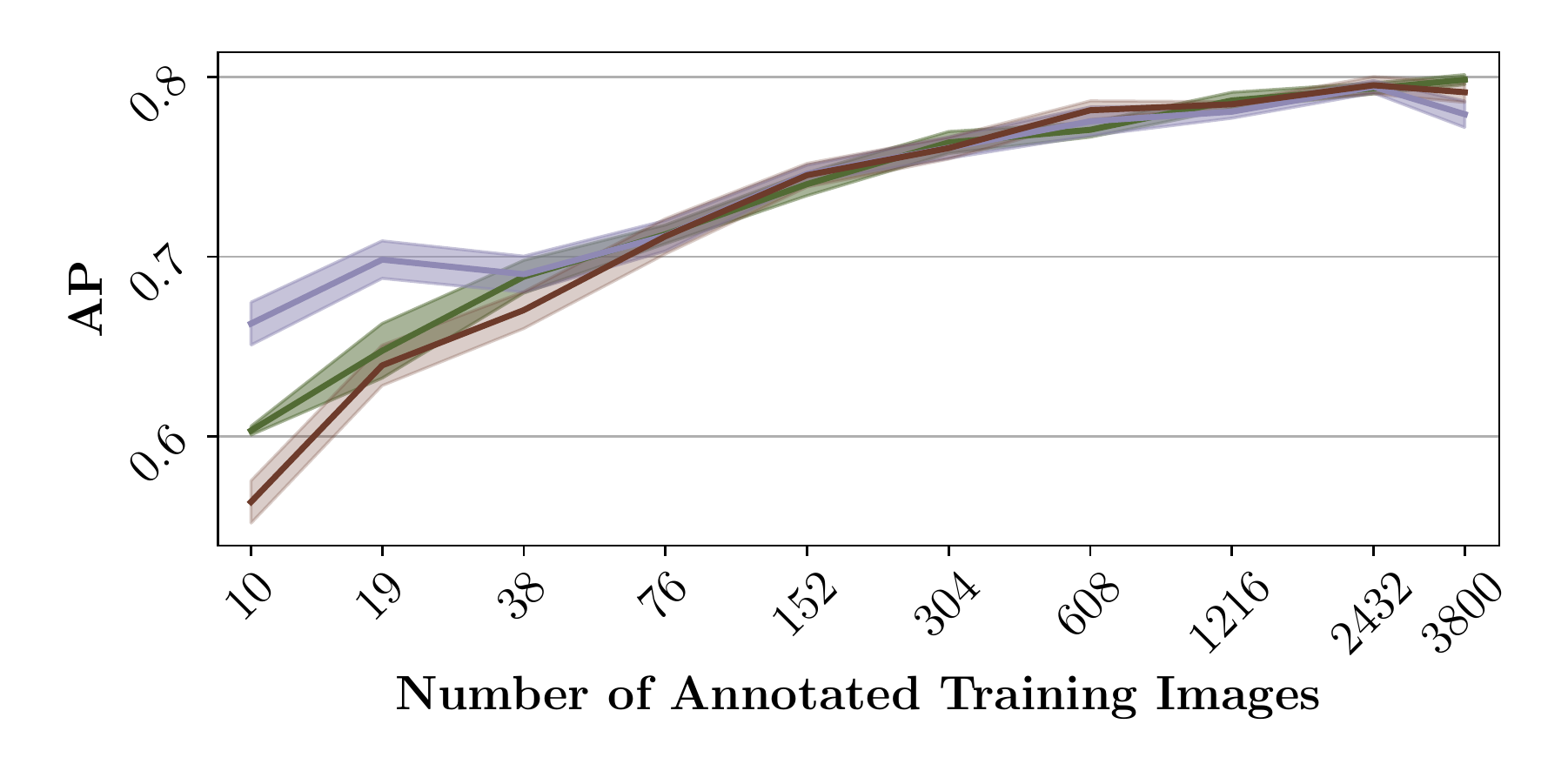}
        \includegraphics[width=.49\linewidth,trim={0.6cm 1.3cm 0.6cm 0.5cm},clip]{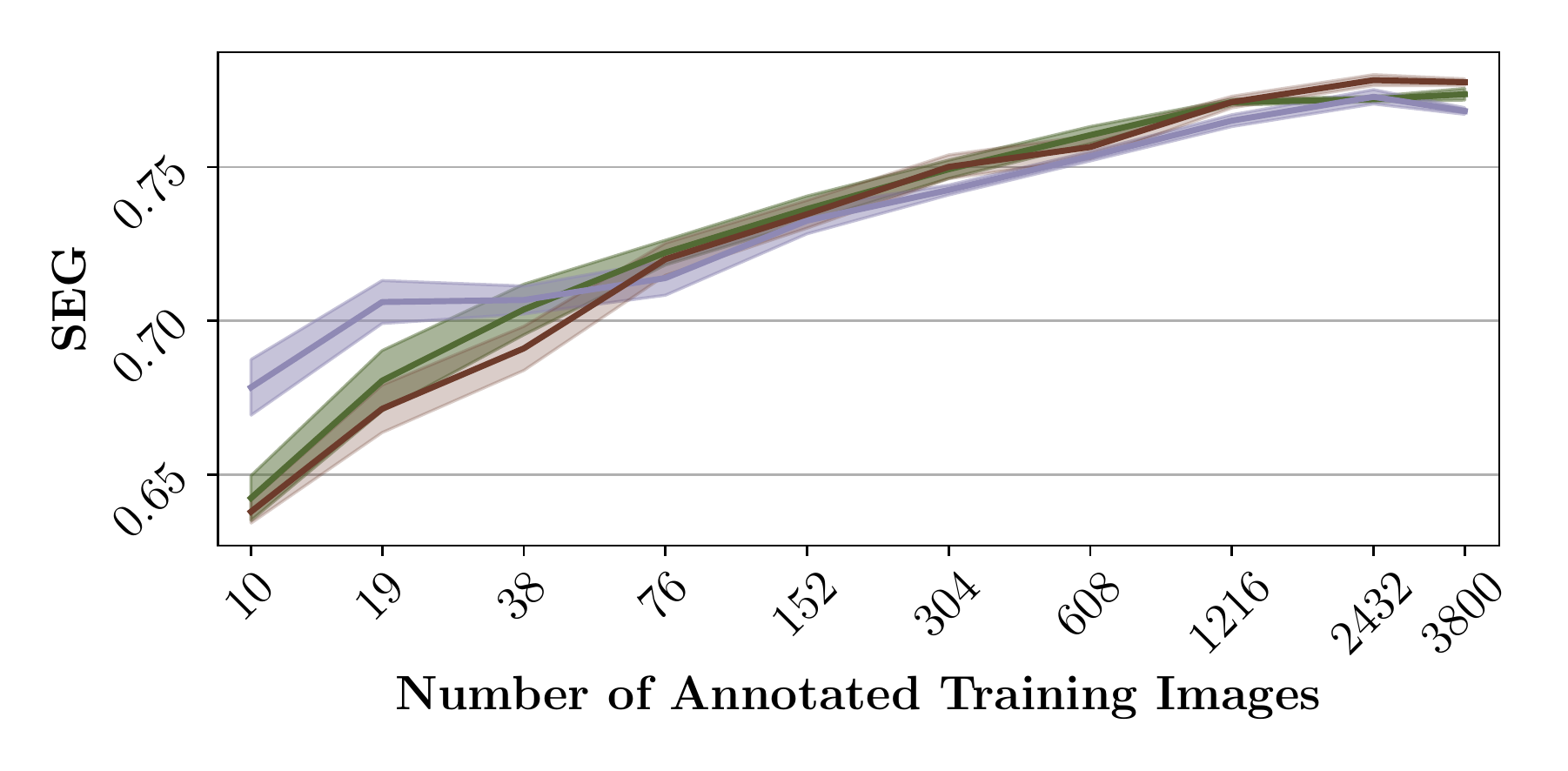}
    \end{minipage}

    \begin{minipage}{.02\linewidth}
    \begin{tikzpicture}
            \draw (0, 0) node[rotate=90] {\textcolor{white}{ws}DSB n0}; 
        \end{tikzpicture}
    \end{minipage}
    \begin{minipage}{.97\linewidth}
        \includegraphics[width=.49\linewidth,trim={0.6cm 0.6cm 0.6cm 0.5cm},clip]{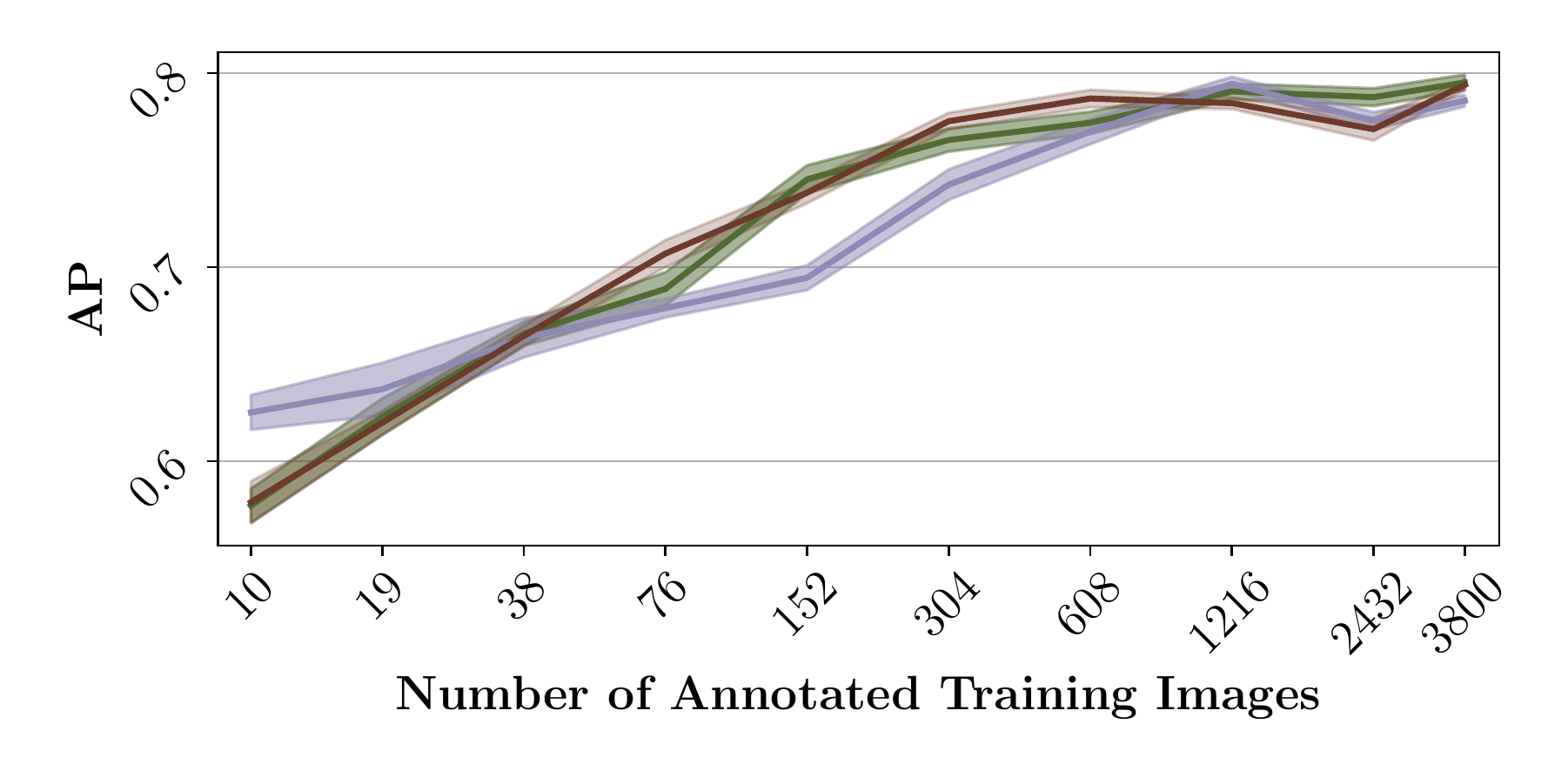}
        \includegraphics[width=.49\linewidth,trim={0.6cm 0.6cm 0.6cm 0.5cm},clip]{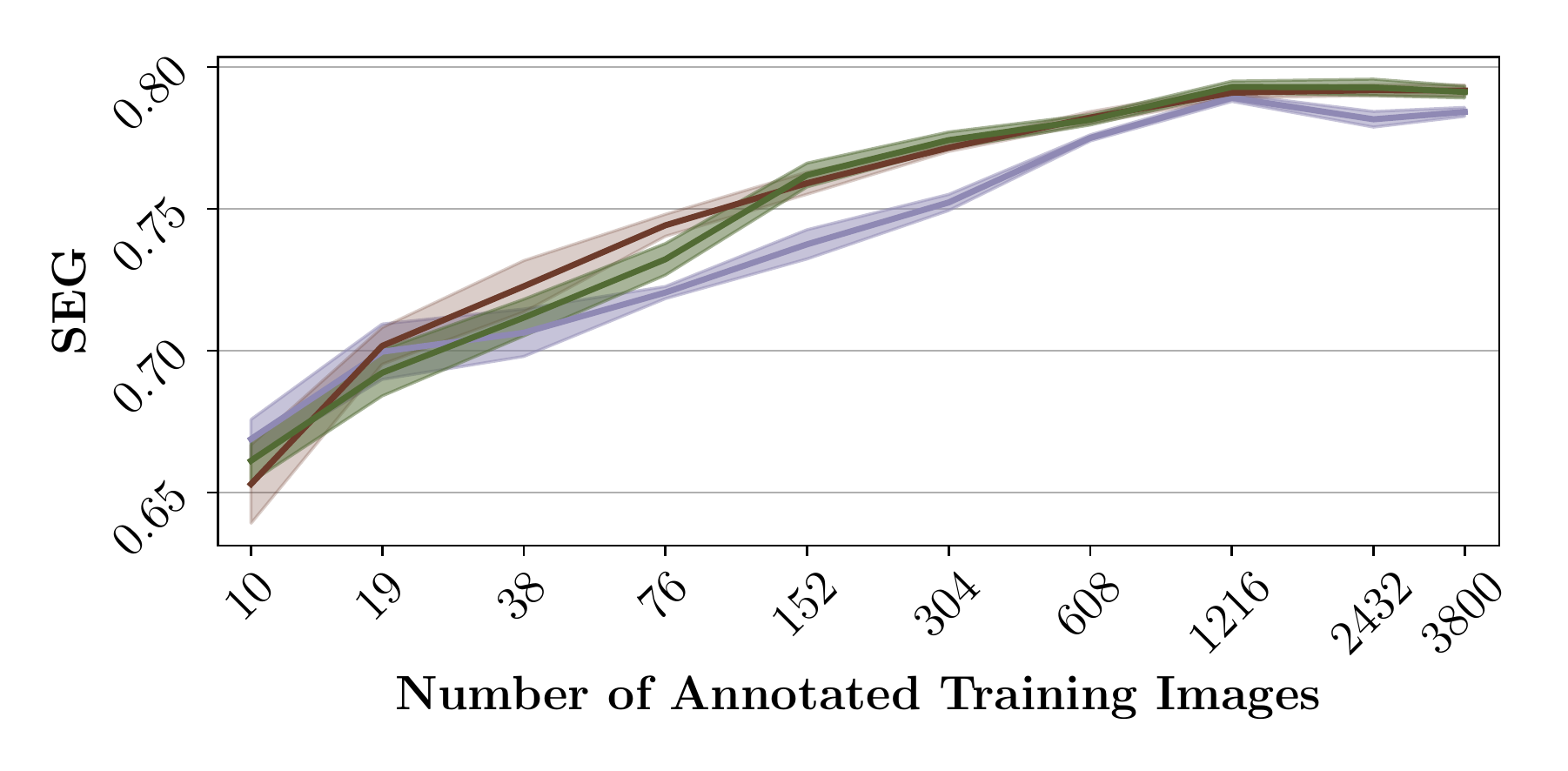}
    \end{minipage}
    
    \caption{Results for DSB n0, n10 and n20, evaluated with Average Precision (AP)~\cite{schmidt2018} and SEG-Score~\cite{ulman2017objective}. 
    \DenoiSeg outperforms both baseline methods, mainly when only limited segmentation 
    ground truth is available.
    Note that the advantage of our proposed method is at least partially compromised when the image data is not noisy (row 3).}
    \label{fig:DSB}
\end{figure}
}

\newcommand\figDeltaNoise{
\begin{figure}[ht]
    \centering
    \begin{minipage}{\linewidth}
        \begin{tikzpicture}
            \draw (0, 0) node[inner sep=0] {\includegraphics[width=\linewidth,trim={0.6cm 0.6cm 0.8cm 0.5cm},clip]{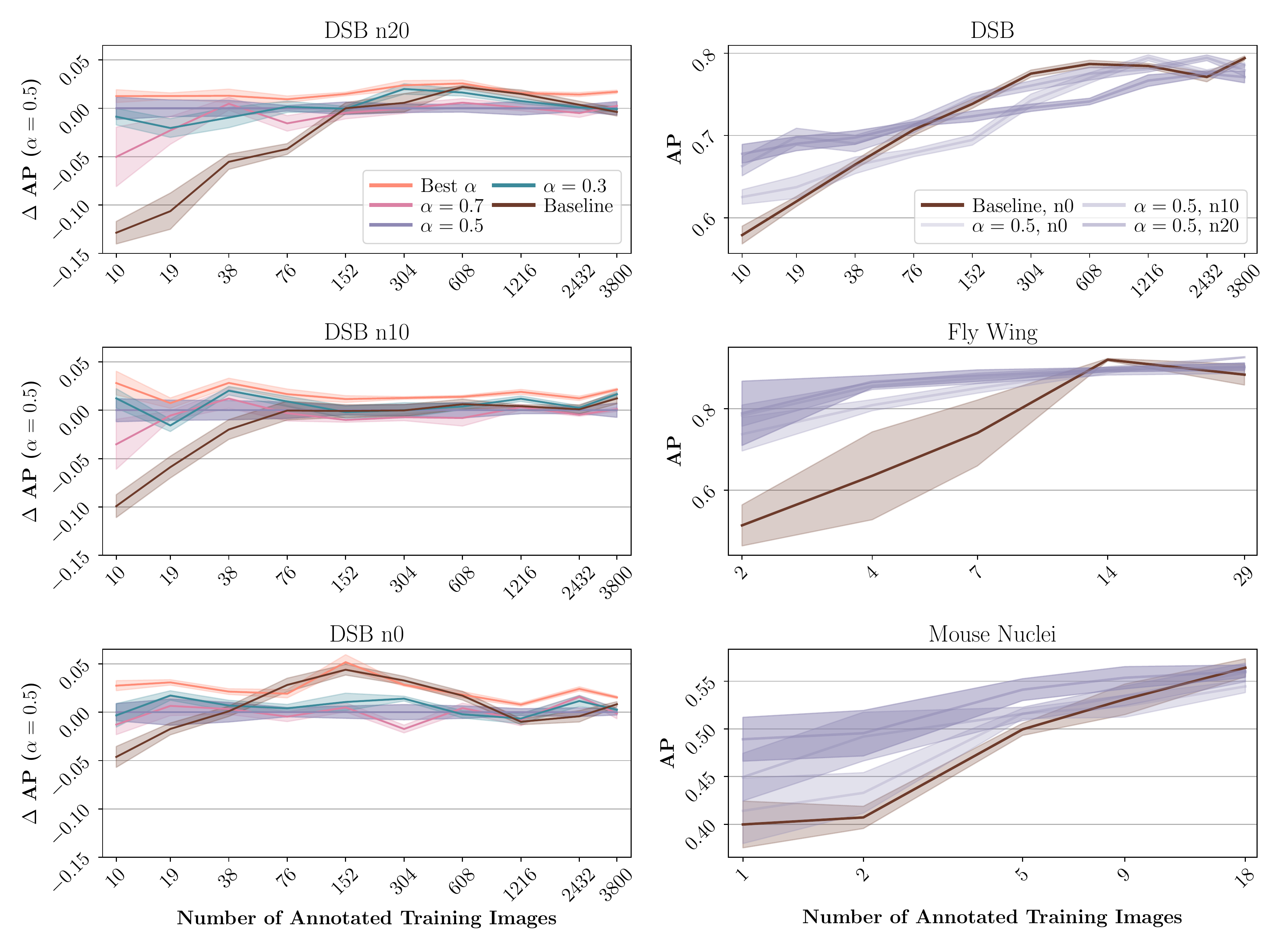}};
            \draw (-5.8, 4.4) node[inner sep=0] {(a)};
            \draw (0.45, 4.4) node[inner sep=0] {(b)};
        \end{tikzpicture}
    \end{minipage}
    
    \caption{In \textbf{(a)}, we show that \DenoiSeg consistently improves results over the baseline for a broad range of hyperparameter $\alpha$ values. The  results come close to what would be achievable by choosing the best possible $\alpha$ (see main text).
    In \textbf{(b)}, we show that adding synthetic noise can lead to improved \DenoiSeg performance. 
    For the DSB, Fly Wing, and Mouse Nuclei data, we compare baseline results with \DenoiSeg results on the same data (n0) and with added synthetic noise (n10 and n20, see main text).
    }
    \label{fig:deltaNoise}
\end{figure}
}

\newcommand\figQualitative{
\begin{figure}[H]
    \centering
    \begin{minipage}{\linewidth}
        \begin{tikzpicture}
            \draw (1.06, 0) node[inner sep=0] {\includegraphics[width=.96\linewidth]{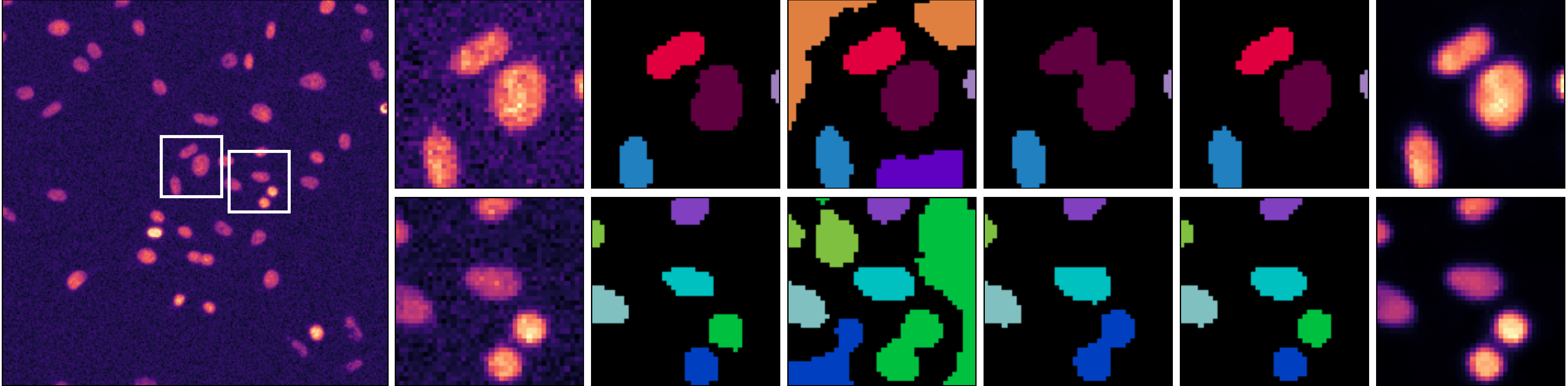}};
            \draw (-5, 0) node[rotate=90] {{\textcolor{white}{g}}DSB n10{\textcolor{white}{g}}}; 
            \draw (-4.3, 1.6) node {Input};
            \draw (-1.11, 1.6) node {{\textcolor{white}{g}}Insets{\textcolor{white}{g}}};
            \draw (0.35, 1.6) node {{\textcolor{white}{g}}GT{\textcolor{white}{g}}};
            \draw (1.8, 1.6) node {{\textcolor{white}{g}}Baseline{\textcolor{white}{g}}};
            \draw (3.25, 1.6) node {Sequent.};
            \draw (5.47, 1.9) node {$\overbrace{\text{\textcolor{white}{blablablablablablab}}}^{\text{\textbf{Ours}}}$};
            \draw (4.75, 1.6) node {Segm.};
            \draw (6.17, 1.6) node {{\textcolor{white}{g}}Denoised{\textcolor{white}{g}}};
            \draw (-3.34, 1.24) node {\textcolor{white}{\textbf{3800 (GT for 10)}}};
        \end{tikzpicture}
        \begin{tikzpicture}
            \draw (1.06, 0) node[inner sep=0] {\includegraphics[width=.96\linewidth]{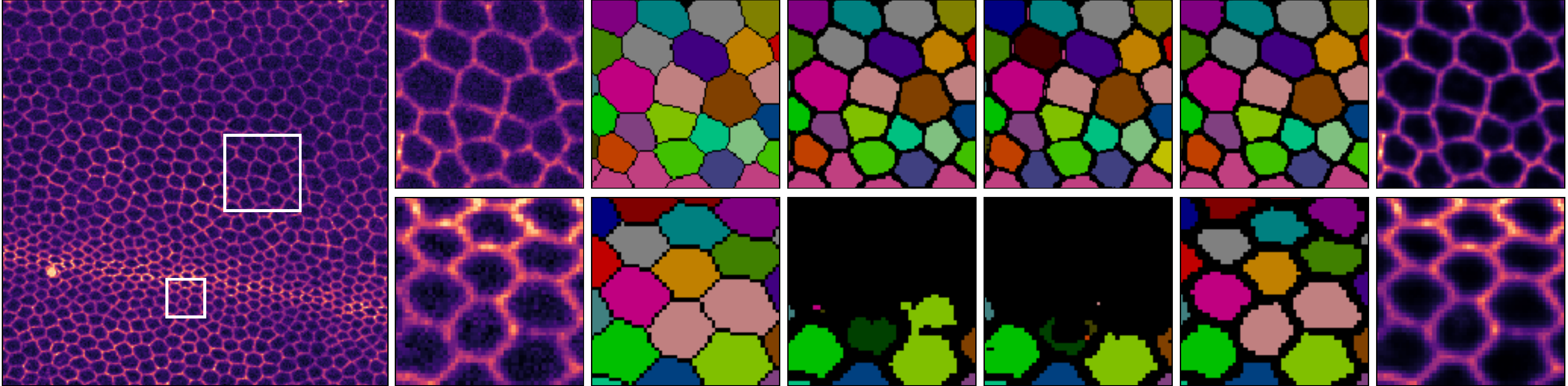}};
            \draw (-5, 0) node[rotate=90] {Fly Wing n10};
            \draw (-3.44, 1.24) node {\textcolor{white}{\textbf{1428 (GT for 2)}}};
        \end{tikzpicture}
        \begin{tikzpicture}
            \draw (1.06, 0) node[inner sep=0] {\includegraphics[width=.96\linewidth]{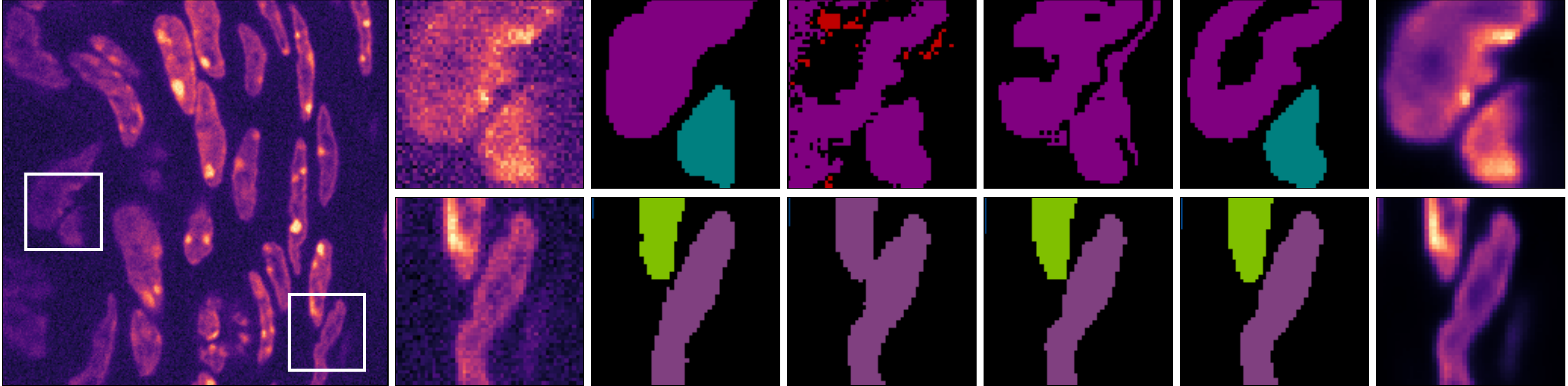}};
            \draw (-5, -0.1) node[rotate=90] {{\textcolor{white}{g}}Mouse Nuclei n10};
            \draw (-3.52, 1.24) node {\textcolor{white}{\textbf{908 (GT for 2)}}};
        \end{tikzpicture}
    \end{minipage}
    
    \caption{Qualitative results on DSB n10 (first row), Fly Wing n10 (second row) and Mouse Nuclei n10 (third row). 
    The first column shows an example test image. Numbers indicate how many noisy input and annotated ground truth (GT) patches were used for training.
    Note that segmentation GT was only available for at most 10 images, accounting for less than 0.27\% of the available raw data.
    Other columns show depicted inset regions, from left to right showing: raw input, segmentation GT, results of two baseline methods, and our \DenoiSeg segmentation and denoising results.}
    \label{fig:qualitative}
\end{figure}
}

\newcommand\tabDenoising{
\begin{table}[h]
\centering  

\begin{tabular}{p{0.8cm}||p{1.75cm}p{1.75cm}| p{1.75cm}p{1.75cm} | p{1.75cm}p{1.75cm}}
\hline
\multicolumn{1}{c||}{} & \multicolumn{2}{c|}{DSB \small{(GT for 10)}} & \multicolumn{2}{c|}{Fly Wing \small{(GT for 2)}} & \multicolumn{2}{c}{Mouse N. \small{(GT for 1)}}         \\ \hline
Noise & $\DenoiSeg$ & $\NoiseVoid$ & $\DenoiSeg$ & $\NoiseVoid$ & $\DenoiSeg$ & $\NoiseVoid$ \\ \hline
n10 & \small{37.57$\pm$0.07} & \small{38.01$\pm$0.05} & \small{33.12$\pm$0.01} & \small{33.16$\pm$0.01} & \small{37.42$\pm$0.10} & \small{37.86$\pm$0.01} \\
n20 & \small{35.38$\pm$0.08} & \small{35.53$\pm$0.02} & \small{30.45$\pm$0.20} & \small{30.72$\pm$0.01} & \small{34.21$\pm$0.19} & \small{34.59$\pm$0.01} \\ \hline
\end{tabular}
\vspace{0.2cm}
\caption{Comparing the denoising performance of \DenoiSeg and \NoiseVoid. 
Mean Peak Signal-to-Noise Ratio values (with $\pm 1$ SEM over 5 runs) are shown. 
Similar tables for \DenoiSeg results when more segmentation GT was available can be found online in the \DenoiSeg-Wiki.
}
\label{tab:denoising}
\end{table}
}

\begin{document}
\title{\DenoiSeg: Joint Denoising and Segmentation}

\titlerunning{\DenoiSeg: Joint Denoising and Segmentation}

\author{Tim-Oliver Buchholz\inst{\ast,1,2} \and
Mangal Prakash\inst{\ast,1,2} \and
Alexander Krull\inst{1,2,3} \and
Florian Jug\inst{1,2,4,\text{\Letter}}}
\authorrunning{T. Buchholz and M. Prakash \etal}
%
\institute{$^1$Center for Systems Biology, Dresden, Germany\\$^2$Max Planck Institute of Molecular Cell Biology and Genetics, Dresden, Germany\\
$^3$Max Planck Institute for Physics of Complex Systems, Dresden, Germany\\
$^4$Fondatione Human Technopole, Milano, Italy\\
\Letter \: \text{jug@mpi-cbg.de}, \text{florian.jug@fht.org}}
\maketitle              
\blfootnote{$^\ast$ Equal contribution (alphabetical order).}

\begin{abstract}
Microscopy image analysis often requires the segmentation of objects, but training data for this task is typically scarce and hard to obtain.
Here we propose \DenoiSeg, a new method that can be trained end-to-end on only a few annotated ground truth segmentations. 
We achieve this by extending \NoiseVoid\cite{krull2019noise2void}, a self-supervised denoising scheme that can be trained on noisy images alone, to also predict dense 3-class segmentations. 
The reason for the success of our method is that segmentation can profit from denoising, especially when performed jointly within the same network.
The network becomes a denoising expert by seeing all available raw data, while  co-learning to segment, even if only a few segmentation labels are available.
This hypothesis is additionally fueled by our observation that the best segmentation results on high quality (very low noise) raw data are obtained when moderate amounts of synthetic noise are added. 
This renders the denoising-task non-trivial and unleashes the desired co-learning effect.
We believe that \DenoiSeg offers a viable way to circumvent the tremendous hunger for high quality training data and effectively enables few-shot learning of dense segmentations.

\end{abstract}

\section{Introduction}
\label{sec:introduction}
The advent of modern microscopy techniques has enabled the routine investigation of 
biological processes at sub-cellular resolution. 
The growing amount of microscopy image data necessitates the development of automated analysis methods, with object segmentation often being one of the desired analyses. 
Over the years, a sheer endless array of methods have been proposed for segmentation~\cite{jug2014bioimage}, but deep learning (DL) based approaches are currently best performing~\cite{caicedo2019evaluation,moen2019deep,razzak2018deep}.
Still, even the best existing methods offer plenty of scope for improvements, motivating further research in this field~\cite{schmidt2018,stringer2020cellpose,hirsch2020patchperpix}.

A trait common to virtually all DL-based segmentation methods is their requirement for tremendous amounts of labeled ground truth (GT) training data, the creation of which is extraordinarily time consuming.
In order to make the most out of a given amount of segmentation training data, data augmentation~\cite{shorten2019survey,zhao2019data} is used in most cases.
Another way to increase the amount of available training data for segmentation is to synthetically generate it, \eg by using Generative Adversarial Networks (GANs)~\cite{ihle2019unsupervised,osokin2017gans,sandfort2019data}.
However, the generated training data needs to capture all statistical properties of the real data and the respective generated labels, thereby making this approach cumbersome in its own right.

For other image processing tasks, such as denoising~\cite{lehtinen2018noise2noise,weigert2018content,buchholz2019cryo}, the annotation problem has been addressed via self-supervised training~\cite{krull2019noise2void,batson2019noise2self,alex2019probabilistic,2019ppn2v}. While previous denoising approaches~\cite{weigert2018content} require pairs of noisy and clean ground truth training images, self-supervised methods can be trained directly on the noisy raw data that is to be denoised.

Very recently, Prakash~\etal~\cite{prakash2019leveraging} demonstrated on various microscopy datasets that self-supervised denoising~\cite{krull2019noise2void} prior to object segmentation leads to greatly improved segmentation results, especially when only small numbers of segmentation GT images are available for training. 
The advantage of this approach stems from the fact that the self-supervised denoising module can be trained on the full body of available microscopy data.
In this way, the subsequent segmentation module receives images that are easier to interpret, leading to an overall gain in segmentation quality even without having a lot of GT data to train on.
In the context of natural images, a similar combination of denoising and segmentation was proposed by Liu~\etal~\cite{liu2017image} and Wang~\etal~\cite{wang2019segmentation}.
However, both methods lean heavily on the availability of paired low- and high-quality image pairs for training their respective denoising module. 
Additionally, their cascaded denoising and segmentation networks make the training comparatively computationally expensive.

\figTeaser

Here, we present \DenoiSeg, a novel training scheme that leverages denoising for object segmentation (see Fig.~\ref{fig:teaser}).
Like Prakash~\etal, we employ the self-supervised \NoiseVoid~\cite{krull2019noise2void} for denoising.
However, while Prakash~\etal rely on two sequential steps for denoising and segmentation, we propose to use a single network to jointly predict the denoised image and the desired object segmentation.
We use a simple \UNet~\cite{RFB15a} architecture, making training fast and accessible on moderately priced consumer hardware.
Our network is trained on noisy microscopy data and requires only a small fraction of images to be annotated with GT segmentations.
We evaluate our method on different datasets and with different amounts of annotated training images.
When only small amounts of annotated training data are available, our method consistently outperforms not only networks trained purely for segmentation~\cite{chen2016dcan,guerrero2018multiclass}, but also the currently best performing training schemes proposed by Prakash~\etal~\cite{prakash2019leveraging}.

\section{Methods}
\label{sec:methods}
We propose to jointly train a single \UNet for segmentation and denoising tasks. While for segmentation only a small amount of annotated GT labels are available, the self-supervised denoising module does benefit from all available raw images.
In the following we will first discuss how these tasks can be addressed separately and then introduce a joint loss function combining the two. 

\subsubsection{Segmentation.}
\label{sec:segmentation}
We see segmentation as a 3-class pixel classification problem~\cite{chen2016dcan,guerrero2018multiclass,prakash2019leveraging} and 
train a \UNet to classify each pixel as foreground, background or border (this yields superior results compared to a simple classification into foreground and background~\cite{schmidt2018}).
Our network uses three output channels to predict each pixel's probability of belonging to the respective class.
We train it using the standard cross-entropy loss, which will be denoted as
$\mathcal{L}_{s}\big( \seg_i,f(\img_i) \big)$, where $\img_i$ is the $i$-th training image, $\seg_i$ is the ground truth 3-class segmentation, and $f(\img_i)$ is the network output.

\subsubsection{Self-Supervised Denoising.}
\label{sec:selfsupervised_denoising}
We use the \NoiseVoid setup described in~\cite{krull2019noise2void} as our self-supervised denoiser of choice.
We extend the above mentioned 3-class segmentation \UNet by adding a forth output channel, which is used for denoising and trained using the \NoiseVoid scheme.
\NoiseVoid uses a Mean Squared Error (MSE) loss, which is calculated over a randomly selected subset of blind spot pixels that are masked in the input image.
Since the method is self-supervised and does not require ground truth, this loss $\mathcal{L}_{d}\big( \img_i,f(\img_i) \big)$ can be calculated as a function of the input image $\img_i$ and the network output~$f(\img_i)$.

\subsubsection{Joint-Loss.}
\label{sec:joint_loss}
To jointly train our network for denoising and segmentation we use a combined loss.
For a given training batch $(\img_1,\seg_1,\dots,\img_m,\seg_m)$ of $m$ images, we assume that GT segmentation is available only for a subset of the raw images.
We define $\seg_i=\boldsymbol{0}$ for images where no segmentation GT is present.
The loss over a batch is calculated as 
\begin{equation}\label{eq:loss}
    \mathcal{L} = \frac{1}{m}\sum_{i=1}^m \alpha \cdot \mathcal{L}_{d}\big( \img_i,f(\img_i) \big) 
    + (1 - \alpha) \cdot \mathcal{L}_{s}\big( \seg_i,f(\img_i) \big),
\end{equation}
where $0\leq \alpha \leq 1$ is a tunable hyperparameter that determines the relative weight of denoising and segmentation during training.
Note that the \NoiseVoid loss is self-supervised, therefore it can be calculated for all raw images in the batch.
The cross-entropy loss however requires GT segmentation and can only be evaluated on a subset of images, where this information is available.
For images where no GT segmentation is available we define $\mathcal{L}_{s}\big( \seg_i=\boldsymbol{0},f(\img_i) \big)=0$.

In the setup described above, setting $\alpha=1$ corresponds to pure \NoiseVoid denoising.
However, setting $\alpha=0$ does not exactly correspond to the vanilla 3-class segmentation, due to two reasons.
Firstly, only some of the images are annotated but in Eq.~\ref{eq:loss} the loss is divided by the constant batch size $m$. This effectively corresponds to a reduced batch size and learning rate, compared to the vanilla method.
Secondly, our method applies \NoiseVoid masking of blind spot pixels in the input image.

\subsubsection{Implementation Details.}
\label{sec:implementation}
Our \DenoiSeg implementation is publicly available\footnote{https://github.com/juglab/DenoiSeg}.
The proposed network produces four output channels corresponding to denoised 
images, foreground, background and border segmentation. 
For all our experiments we use a \UNet architecture of depth $4$, 
convolution kernel size of $3$, a linear activation function in the last layer, 
$32$ initial feature maps, and batch normalization during training. All 
networks are trained for $200$ epochs with an initial learning rate of $0.0004$.
The learning rate is reduced if the validation loss is not decreasing over ten epochs. 
For training we use $8$-fold data augmentation by adding $90^\circ$ rotated and flipped versions of all images.

\section{Experiments and Results}
\label{sec:results}
We use three publicly available datasets for which GT annotations are available (data available at \DenoiSeg-Wiki\footnote{https://github.com/juglab/DenoiSeg/wiki}).
For each dataset we generate noisy versions by adding pixel-wise independent Gaussian noise with zero-mean and standard deviations of $10$ and $20$. 
The dataset names are extended by n0, n10, and n20 to indicate the respective additional noise. 
For network training, patches of size $128 \times 128$ are extracted and randomly split into training ($85\%$) and validation ($15\%$) sets. 

\begin{itemize}
    \item \textbf{DSB.} From the Kaggle 2018 Data Science Bowl challenge, we take the same images as used by ~\cite{prakash2019leveraging}.
                            The training and validation sets consist of $3800$ and $670$ patches respectively, while the test set counts $50$ images.
    \item \textbf{Fly Wing.} This dataset from our collaborators consist of $1428$ training and $252$ validation patches of a membrane labeled fly wing. 
                            The test set is comprised of $50$ additional images.
    \item \textbf{Mouse Nuclei.} Finally, we choose a challenging dataset depicting diverse and non-uniformly clustered nuclei in the mouse skull, consisting of $908$ training and 160 validation patches.  The test set counts $67$ additional images.
\end{itemize}

\figQualitative

For each dataset, we train \DenoiSeg and compare it to two different competing methods: \DenoiSeg trained purely for segmentation with $\alpha = 0$ (referred to as \textit{Baseline}), and a sequential scheme based on~\cite{prakash2019leveraging} that first trains a denoiser and then the aforementioned baseline (referred to as \textit{Sequential}). 
We chose our network with $\alpha = 0$ as baseline to mitigate the effect of batch normalization on the learning rate as described in Section~\ref{sec:methods}. 
A comparison of our baseline to a vanilla 3-class \UNet with the same hyperparameters leads to very similar results and can be found in the supplementary material.
Furthermore, we investigate \DenoiSeg performance when trained with different amounts of available GT segmentation images. 
This is done by picking random subsets of various sizes from the available GT annotations.
Note that the self-supervised denoising task still has access to all raw input images. A qualitative comparison of \DenoiSeg results with other baselines (see Figure~\ref{fig:qualitative}) indicates the effectiveness of our method.

As evaluation metrics, we use Average Precision (AP)~\cite{everingham2010pascal} and SEG~\cite{ulman2017objective} scores.
The AP metric measures both instance detection and segmentation accuracy while SEG captures the degree of overlap between instance segmentations and GT.
To compute the scores, the predicted foreground channel is thresholded and connected components are interpreted as instance segmentations. 
The threshold values are optimized for each measure on the validation data. 
All conducted experiments were repeated $5$ times and the mean scores along with $\pm 1$ standard error of the mean are reported in Figure~\ref{fig:DSB}.

\subsubsection{Performance with Varying Quantities of GT Data and Noise.}
\figDSB
Figure~\ref{fig:DSB} shows the results of \DenoiSeg with $\alpha = 0.5$ (equally weighting denoising and segmentation losses) for DSB n0, n10 and n20 datasets. 
For low numbers of GT training images, \DenoiSeg outperforms all other methods. 
Figures for the other two datasets can be found in the supplementary material. 
Results for all performed experiments showing overall similar trends and can be found on the \DenoiSeg-Wiki.

\subsubsection{Importance of $\alpha$.}
\figDeltaNoise
We further investigated the sensitivity of our results to the hyperparameter $\alpha$. 
In Figure~\ref{fig:deltaNoise}(a) we look at the difference in resulting AP ($\Delta$) when instead of $\alpha=0.5$ we use values of $\alpha=0.3$ and $\alpha=0.7$. Additionally we also compare to the Baseline and results that use (the a priori unknown) best $\alpha$. 
The best $\alpha$ for each trained network is found by a grid search for $\alpha \in \{0.1, 0.2, \dots, 0.9\}$. 
Figure~\ref{fig:deltaNoise}(a) shows that our proposed method is extraordinarily robust with respect to the choice of $\alpha$. 
Results for the other datasets showing similar trends can be found in the supplementary material. 

\subsubsection{Noisy Inputs Lead to Elevated Segmentation Performance.}
Here we want to elaborate on the interesting observation we made in Figure~\ref{fig:DSB}: when additional noise is synthetically added to the raw data, the segmentation performance reaches higher AP and SEG scores, even though  segmentation should be more difficult in the presence of noise.
We investigate this phenomenon in Figure~\ref{fig:deltaNoise}(b).
We believe that in the absence of noise the denoising task can be solved trivially, preventing the regularizing effect that allows \DenoiSeg to cope with small amounts of training data.

\sloppy

\subsubsection{Evaluation of Denoising Performance.}
Although we are not training \DenoiSeg networks for their denoising capabilities, it is interesting to know how their denoising predictions compare to dedicated denoising networks.
Table~\ref{tab:denoising} compares our denoising results with results obtained by \NoiseVoid~\cite{krull2019noise2void}. It can be seen that co-learning segmentation is only marginally impeding the network's ability to denoise its inputs.

\fussy

\tabDenoising

\section{Discussion}
\label{sec:discussion}
Here we have shown that
$(i)$~joint segmentation and self-supervised denoising leads to improved segmentation quality when only limited amounts of segmentation ground truth is available (Figures~\ref{fig:qualitative} and~\ref{fig:DSB}),
$(ii)$~the hyperparameter $\alpha$ is modulating the quality of segmentation results but leads to similarly good solutions for a broad range of values,
and $(iii)$~results on input data that are subject to a certain amount of intrinsic or synthetically added noise lead to better segmentations than \DenoiSeg trained on essentially noise-free raw data.

We reason that the success of our proposed method originates from the fact that similar \say{skills} are required for denoising and segmentation. 
The segmentation task can profit from denoising, and compared to~\cite{prakash2019leveraging}, performs even better when jointly trained within the same network.
When a low number of annotated images are available, denoising is guiding the training and the features learned from this task, in turn, facilitate segmentation.

We believe that \DenoiSeg offers a viable way to enable few-shot learning of dense segmentations and can therefore be applied in cases where other methods cannot.
We also show that the amount of required training data can be so little, even ad-hoc label generation by human users is a valid possibility, expanding the practical applicability of our proposed method manyfold.

\subsubsection*{Acknowledgments.}
\label{sec:acknowledgments}
The authors would like to acknowledge Romina Piscitello-Gomez and Suzanne Eaton from MPI-CBG for fly wing data, Diana Afonso and Jacqueline Tabler from MPI-CBG for mouse nuclei data and the Scientific Computing Facility at MPI-CBG 
for giving us access to their HPC cluster.

\newpage

%

\bibliographystyle{splncs04}
\bibliography{refs}
\end{document}